\documentclass{article}
\usepackage{cite}
\usepackage{amsmath,amssymb,amsfonts}
\usepackage{algorithmic}
\usepackage{graphicx}
\usepackage{textcomp}
\usepackage{booktabs}
\usepackage{multirow}
\usepackage{algorithm}
\usepackage{adjustbox}
\usepackage{algorithm}
\usepackage{algorithmic}
\usepackage{bm}
\usepackage{geometry}

\title{A Transformer-in-Transformer Network Utilizing Knowledge Distillation for Image Recognition}

\author{
Dewan Tauhid Rahman\thanks{Department of Computer Science, University of Miami, Florida 33156, USA. Email: dxr1367@miami.edu}, 
Yeahia Sarker\thanks{Department of Mechatronics Engineering, Rajshahi University of Engineering \& Technology (RUET), Rajshahi 6200, Bangladesh. Email: yeahia.ruet@gmail.com}, 
Antar Mazumder\thanks{Department of Mechatronics Engineering, Rajshahi University of Engineering \& Technology (RUET), Rajshahi 6200, Bangladesh. Email: antar.mte@ieee.org}, 
Md. Shamim Anower\thanks{Department of Electrical \& Electronic Engineering, Rajshahi University of Engineering \& Technology (RUET), Rajshahi 6200, Bangladesh. Email: md.shamimanower@yahoo.com}
}

\begin{document}
\maketitle

\begin{abstract}
This paper presents a novel knowledge distillation neural architecture leveraging efficient transformer networks for effective image classification. Natural images display intricate arrangements encompassing numerous extraneous elements. Vision transformers utilize localized patches to compute attention. However, exclusive dependence on patch segmentation proves inadequate in sufficiently encompassing the comprehensive nature of the image. To address this issue, we have proposed an inner-outer transformer-based architecture, which gives attention to the global and local aspects of the image. Moreover, The training of transformer models poses significant challenges due to their demanding resource, time, and data requirements. To tackle this, we integrate knowledge distillation into the architecture, enabling efficient learning. Leveraging insights from a larger teacher model, our approach enhances learning efficiency and effectiveness. Significantly, the transformer-in-transformer network acquires lightweight characteristics by means of distillation conducted within the feature extraction layer. Our featured network's robustness is established through substantial experimentation on the MNIST, CIFAR10, and CIFAR100 datasets, demonstrating commendable top-1 and top-5 accuracy. The conducted ablative analysis comprehensively validates the effectiveness of the chosen parameters and settings, showcasing their superiority against contemporary methodologies. Remarkably, the proposed Transformer-in-Transformer Network (TITN) model achieves impressive performance milestones across various datasets: securing the highest top-1 accuracy of 74.71\% and a top-5 accuracy of 92.28\% for the CIFAR100 dataset, attaining an unparalleled top-1 accuracy of 92.03\% and top-5 accuracy of 99.80\% for the CIFAR-10 dataset, and registering an exceptional top-1 accuracy of 99.56\% for the MNIST dataset.
\end{abstract}


\section{Introduction}
\label{sec:introduction}
The advent of machine learning approaches throughout the years such as regression \cite{reg1}, instance-based statical analysis ~\cite{ins1}, regularization ~\cite{regu1}, decision trees~\cite{dt1}, bayesian ~\cite{bay1}, clustering ~\cite{clus1}, and the recent surge in the development of artificial neural networks has yielded significant progress across a diverse spectrum of computational tasks, marking notable evolutionary strides in the field. Among these fields of practice, computer vision holds unparalleled significance. Various approaches have been applied to image classification, an important computer vision problem over the years; nevertheless, the methodology experienced a paradigm change with the use of Artificial Neural Networks (ANNs), especially the Convolutional Neural Networks (CNNs), the now most prominent approach for image classification~\cite{cnngood}. Prior to CNN, typical ANNs often failed to maintain image information if dimensionality was reduced. Feature optimization was another critical task that lacked balance. It is the emergence of CNNs that made it possible for substantial dimensionality reduction with almost no loss of information. Additionally, the feature optimization capabilities of CNN~\cite{featurecnn} greatly outperformed its predecessors. In fact, CNNs have been the most extensively used deep neural architectures in computer vision over the previous decade due to their exceptional performance, capability, and adaptability.

In general, two CNN approaches are widely implemented in image classification. The standard approach has the same convolution kernel storing coupled channel and spatial correlations while the other approach known as the "depth-wise CNN" decouples them~\cite{zhao2021battle}. Multiple research showed that the latter approach outperformed the standard approach in terms of both accuracy and efficiency~\cite{tan2019efficientnet}. However, as the complexity of such models rose, the corresponding growth of computational load and extensive storage made it more challenging for these models to be applied in real-time applications such as regular video surveillance~\cite{slowsurv}, aerial surveillance~\cite{kim2021rgdinet}, hobby-crafts, human-robot interaction~\cite{melinte2020facial}, autonomous mobile robots, and self-driving vehicles~\cite{slowself}. On the other hand, as the modern world applications became more and more feature-demanding the functionalities required by such applications required even more complex models resulting in a tension between requirement and performance. Thus there was and still is a constant need for models with optimal depth and minimal execution time with versatile system-scale compatibility. Albeit, CNNs and other ANNs have been greatly optimized but eventually these approaches tend to hit a threshold beyond which they become unfeasible due to either extensive depth complexity or poor performance due to not having adequate depth. Thus, considering the present world's needs, an alternative method to break down tasks into multiple optimal models that cooperate with each other is in demand. Research suggested that sequential information could be a valuable asset for optimizing in image classification process~\cite{sequence}. However, such information is difficult to obtain from the standard CNN or ANN method. For such reasons, when it comes to obtaining sequential information and exhibiting the connections between various model properties another recent technique known as the vision transformer has been gaining traction in academics which can also replicate the decoupling approach~\cite{zhao2021battle}.

Vaswani \textit{et al.} described a transformer as a type of neural network that uses a self-attention process to handle long-term dependencies while solving transduction issues of sequence-to-sequence processes~\cite{vaswani2017attention}. So far, the most well-known transformers were the Bidirectional Encoder Representations from Transformers and Generative Pre-trained Transformer 3 ~\cite{brown2020language} models. However, both of these models were primarily applied to Natural Language Processing (NLP) applications showing exceptional results. Recently academics focused on using the technique for visual tasks, especially in order to obtain sequential information and to replicate the decoupling approach found in depth-based CNNs ~\cite{zhao2021battle}. However, there are specific differences between approaches taken for NLP and those for computer vision, such as the semantic gap between input images and ground-truth labels. In the case of NLP, there is a semantic gap between input and ground-truth labels, whereas the gap is absent in the case of machine vision applications, which made it challenging to apply traditional transformers for machine vision applications. Again, despite the rising promises of vision transformers for computer vision applications it is very difficult to use deep models on mobile devices and embedded systems because they have very little processing power and memory. Implementation of such models is also challenging for real-time applications where execution speed is a critical concern, which is true for almost every big model. 
For years, several approaches were made to harness the self-attentive features of transformers for machine vision applications such as image detection and classification problems. Furthermore, for visual tasks, a number of academics have looked at how to express sequence information from multiple data sets using transformer structures. Some went on to determine the feasibility of such approaches. Self-attention mechanisms in non-local networks, for example,  have been studied by Wang \textit{et al.} for video and image recognition showing potential results~\cite{wang2018non}. A transformer encoder-decoder design dubbed Detection Transformer or shortly DETR was used by Carion \textit{et al.} to tackle the object detection issue where DETR outperformed Region-based Convolutional Neural Network(R-CNN) while detecting large objects ~\cite{carion2020end}. Parmer \textit{et al.} proposed an image transformer model that could provide more robust receptive fields than contemporary CNNs~\cite{image2018image}.  Chen \textit{et al.} pioneered the use of self-supervised pre-training for image recognition on a pure transformer model without convolution that exhibited a 72\% top-1 accuracy~\cite{chen2020generative}. A very recent approach by Han \textit{et al.} explored a transformer in transformer method that resulted in a top-1 accuracy of 81.5\%~\cite{han2021transformer}. Such approaches, however, lost feasibility to significant extents when in the case of limited processing and memory capacity devices such as mobile devices and embedded systems.
It was initially suggested by~\cite{urban2016deep} to reduce the size of a big model or ensemble of models in order to train a smaller model without significantly reducing accuracy. Thus, a semi-supervised teacher-student model was needed which was complemented by knowledge distillation.

Hinton \textit{et al.} described knowledge distillation as the process of learning a smaller model from a larger one~\cite{hinton2015distilling}. Typically, a teacher supervises a small student model in knowledge distillation. The student model copies the instructor model in order to achieve competitive or even better performance in comparison to their peers. There is a major challenge in transferring information from a big teacher model to a smaller student model. As a remedy to the aforementioned issue, we can look into model compression, which was inspired by knowledge distillation, which aims to minimize the training burden of deep models by distilling data from a huge dataset into an even smaller one, or “dataset distillation"~\cite{wang2018dataset}. 

Model compression uses knowledge distillation in a method that's analogous to how people learn. Recent knowledge distillation approaches have been inspired by this and extended to teacher-student learning~\cite{hinton2015distilling}, mutual learning~\cite{zhang2020reliable}, assistant teaching~\cite{mirzadeh2020improved}, lifelong learning~\cite{zhai2019lifelong}, and self-learning~\cite{Yuan2019RevisitKD}. Compressing deep neural networks is the primary focus of most knowledge distillation expansions. The utilization of lightweight student networks finds pertinence in a range of domains encompassing image recognition, audio analysis, and NLP, showcasing their applicative depth and significance. Aside from adversarial assaults, data augmentation, privacy and security, and knowledge transfer from one model to the next in knowledge distillation are also possible extensions~\cite{papernot2016distillation}. However, among all these potentials there are also some very specific limitations such as extensive training time for transformers and inadequate image classification results when knowledge distillation is merged with the transformer. These methods requires extensive computational power but lack poor feature extraction capabilities due to weak representation learning process. Our transformer-in-transformer approach minimizes the need for higher training resources as well as focuses on a better feature learning system. The proposed semi-supervised effectively divides the datasets into smaller chunks through a data distillation process in order to mitigate the necessity of large-scale data-driven models.

\begin{itemize}
    \item The featured transformer-in-transformer architecture with knowledge distillation enables simultaneous exploration of local and global features in natural images, facilitating faster learning from the teacher model while minimizing resource requirements.
    
    \item A novel loss function has been featured leveraging cutmix loss function with the base loss. In order to use the potential of regional dropout, we have integrated cutmix loss as a part of our hybrid loss function for better localization of spatial features. Substantial results have proved that our criterion can strongly discriminate weakly labelled data during the training process.

    \item Rigorous experiments conducted across MNIST, CIFAR10, and CIFAR100 datasets substantiate the effectiveness of the proposed approach. The empirical \textbf{evaluation} reveals noteworthy improvements in execution speed and accuracy, firmly establishing a performance benchmark.

\end{itemize}

The remaining paper is organized into four additional sections. Section II provides the previous works related to knowledge distillation and other similar approaches. Section III delves into the methodology adopted in this research including descriptions of the datasets used, preprocessing, and suggested model followed by Section IV where the experimental settings and the results are discussed. Finally, the concluding remarks are added in section V.

\section{Related Works}

\subsection{Vision Transformer}

Dosovitskiy \textit{et al.} introduced the vision transformer or more popularly, the ViT~\cite{dosovitskiy2020image}, which facilitated the use of transformer-based models for vision problems by breaking down the input image into several small patches, termed visual sequences, enabling the natural calculation of attention between any two image patches. Subsequently, researchers in~\cite{touvron2021training} explored data-efficient training and distillation to enhance ViT's performance on the ImageNet benchmark, achieving a top-1 accuracy of 81.8\% through extensive experiments, which was comparable to state-of-the-art convolutional networks. Recent surveys indicate a growing adoption of transformer architectures in computer vision tasks over the last few years, including image recognition~\cite{han2022survey}, object detection~\cite{zhu2020deformable}, and segmentation~\cite{zheng2021rethinking}, as well as other tasks. However, as execution speed increases, maintaining good accuracy becomes a growing concern. As a remedy to such issues, Bucilua \textit{et al.} developed a model compression approach that enables the transfer of knowledge from a large model or an ensemble of models to a smaller model, mitigating the accuracy drop typically associated with model compression~\cite{bucilua2006model}.


\subsection{Attention Mechanism}
Attention mechanism has been widely used in several transformer-based models. It has been the core part to learn long term information in terms of feature extraction. Attention in transformer model has been used in many tasks including object recognition \cite{heo2022occlusion}, image classification \cite{wang2021not}, image super-resolution \cite{lu2022transformer}, image translation etc. Transformers utilize scaled dot-product and multi-head attention mechanisms, enhancing computational efficiency and capturing more complex data relationships \cite{fateh2024advancing}. These mechanisms have been important in allowing transformers to outperform traditional convolutional networks in various visual learning tasks. Among all attention mechanisms, channel attention modules focus on enhancing significant feature channels while suppressing less relevant ones, primarily introduced by SENet \cite{hu2018squeeze}. They use a two-step process: squeezing global spatial information and exciting inter-channel dependencies. On the other hand, spatial attention mechanisms target specific regions within an image, highlighting important areas and diminishing background noise \cite{zhu2019empirical}. They generate spatial attention maps that assign weights to different image locations, improving feature expression. Another approach, branch attention mechanisms dynamically select and emphasize different network branches based on input data, optimizing feature learning \cite{fukui2019attention}. However, all of these methods are not lightweight and do not cover deeper and longer feature dependencies.

\subsection{Knowledge Distillation}

Most of the new ideas for distilling knowledge focus on compressing very large neural networks. The lightweight student networks can be used in applications such as visual recognition, speech recognition, and natural language processing, and they can be set up quickly. It can also be used for other things, like adversarial attacks~\cite{papernot2016distillation}, adding data~\cite{Yuan2019RevisitKD}, protecting data privacy and security, and more. The idea of knowledge distillation for model compression has been used to compress the training data, which is called dataset distillation. This process moves the knowledge from a large dataset into a small dataset to make it easier for deep models to train~\cite{bohdal2020flexible}.
In a recent study, Cheng \textit{et al.} measured how many visual concepts were extracted from the intermediate layers of a deep neural network in order to show how knowledge was boiled down~\cite{cheng2020explaining}. Risk bound, data efficiency, and imperfect teachers all played a part in how knowledge was distilled on a wide neural network~\cite{ji2020knowledge}. Knowledge distillation had also been used to make labels smoother, to check the accuracy of the teacher, and to figure out what the best shape for the output layer should be~\cite{tang2020understanding}. However, a recent study by Cho \textit{et al.} performed extensive experiments to see if knowledge distillation worked but results from the experiments suggested otherwise ~\cite{yang2020knowledge}. It was theorized that the poor performance of knowledge distillation was linked to the fact that a bigger model may not be a better teacher because it, albeit being a larger model, might not have enough space for performing all the intended tasks~\cite{mirzadeh2020improved}.


\section{Methodology}
Transformer-based architecture requires more training data than convolutional-based models. Thus, their performance drops down on small-scale datasets. In order to utilize the effectiveness of the feature extraction of the transformer on small-scale data, we proposed a new variant of transformer architecture using a knowledge distillation procedure that works on various benchmark image datasets. In this section, we give a brief discussion of the preliminaries as well as define the working procedure of our proposed method. 

\subsection{Preliminaries}

\subsubsection{Multi-head Self Attention Mechanism}
Multi-head self-attention transforms the input into three parts, i.e. K (key), Q (query) \& V(value).  \textit{Q}, K and V are split into multiple numbers of heads and the scaled dot-product is then applied in parallel. The output values of each head are added and then a linear layer is used to project the final output. scaled dot-product attention can be defined as follows \cite{koizumi2020speech}:

\begin{equation}
    \textbf{Attention(Q, K, V)} = softmax(\frac{QK^T}{\sqrt{d_k}})V
\end{equation}
where $\sqrt{d_k}$ is the dimension of the key vector $k$ and query vector $q$ . 
So, for Multi-head self-attention can be written as : 
\begin{equation}
    \textbf{MultiHead(Q, K, V)} = Concat(head_1, ..., head_h)W^O
\end{equation}
where, 
\begin{equation}
    \mathbf{head}_{i} = Attention(Q W^Q_i, K W^K_i, V W^V_i)
\end{equation}
\subsubsection{Multi-layer Perceptron Block}

MLP block is applied between self-attention layers for robust feature transformation :

\par

\begin{equation}
    \mathbf{MLP}(X) = \mathbf{FC}(\sigma(\mathbf{FC}(\mathcal{X}))) 
\end{equation}

\begin{equation}
    \mathbf{FC}(X) = \mathbf{X} \odot \mathbf{W} + \mathbf{b} 
\end{equation}

where, \(X\) is the input image, $\sigma(.)$ is the activation function, such as GELU \cite{hendrycks2016gaussian}, and $W$ and $b$ are the weight and bias terms of the fully-connected layer, respectively.
\par
Layer Normalization is an essential part of stable training and faster convergence of transformers \cite{xu2019understanding}.  LN is applied over each sample using the following equation: 
\par
\begin{equation}
    \mathcal{LN}(x) = \frac{x - \mu}{\delta} \circ \gamma + \beta
\end{equation}
where $\mu \in R$, $\delta \in R$ are the feature's mean \& standard deviation, respectively. $\circ$ is the element-wise dot, and $\gamma \in R^d$, $\beta \in R^d$ are affine transform parameters that are learnable.

\subsection{Neural Architecture of TITN}
First, an image input is split into bigger patches, each with a resolution of (p, p), where p is the patch size. For these larger patches, there is a distinct, parallel data flow. Class tokens, distillation tokens, and patch positional embeddings are all mixed together in one path. With a resolution of (m, m), where m is the desired smaller patch size, these larger patches are further divided into smaller ones in the other direction. The result is pixel-level patches, which are created by combining these smaller patches with patch-positional embeddings. These pixel-level patches are sent into the inner transformer blocks, and the output is added to the input in a residual manner \cite{gou2021knowledge}. 

\begin{figure}[t]
    \centering
    \includegraphics[width=0.4\textwidth]{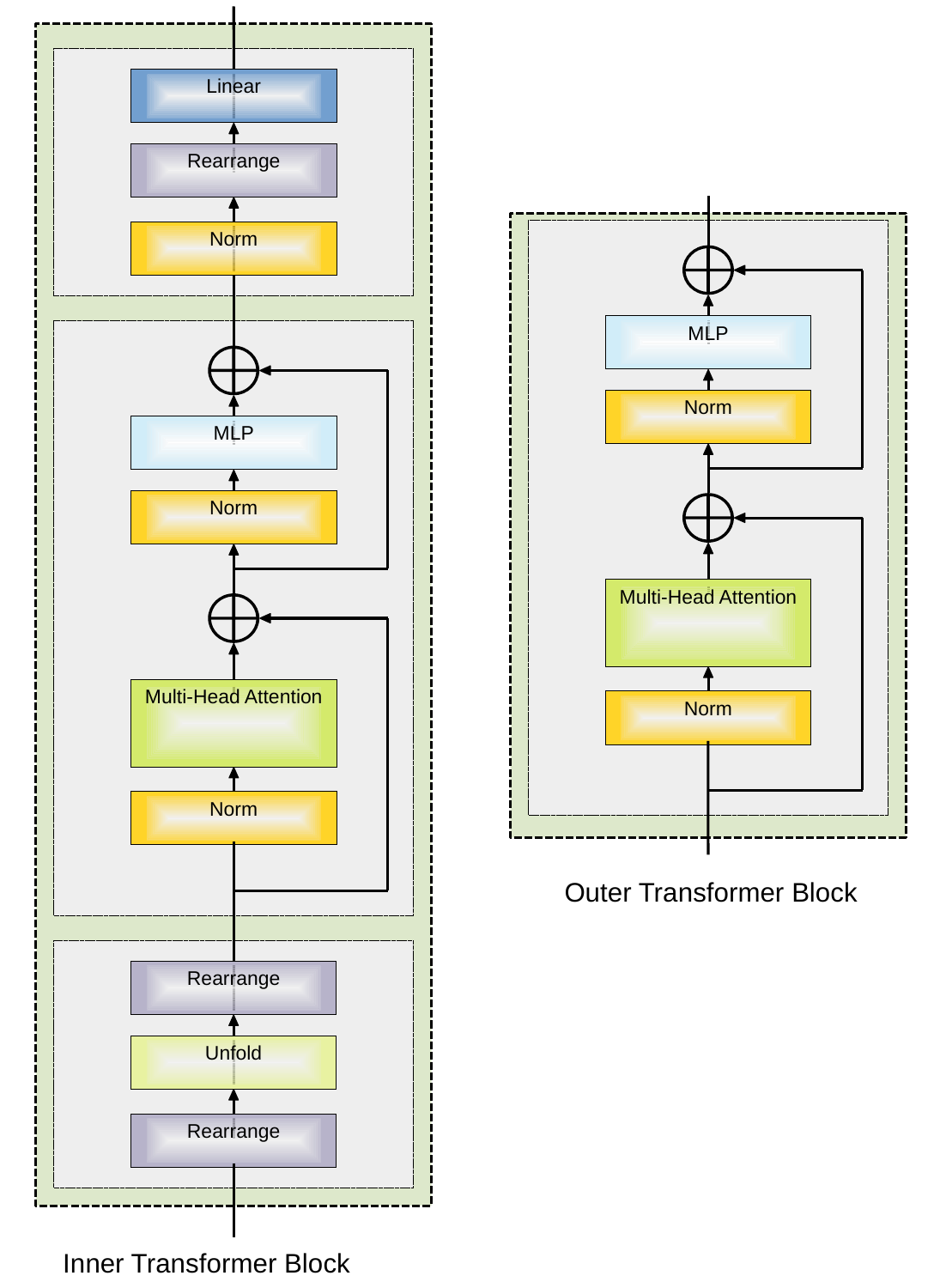}
    \caption{Inner \& outer transformer block of the proposed transformer-in-transformer network. Here, both blocks perform sequentially to generate class and distillation tokens. The inner transformer block rearranges the layer to be fed for the architecture.}
    \label{fig:transformer_block}
\end{figure}

\begin{figure}[t!]
    \centering
    \includegraphics[width = 0.40\textwidth]{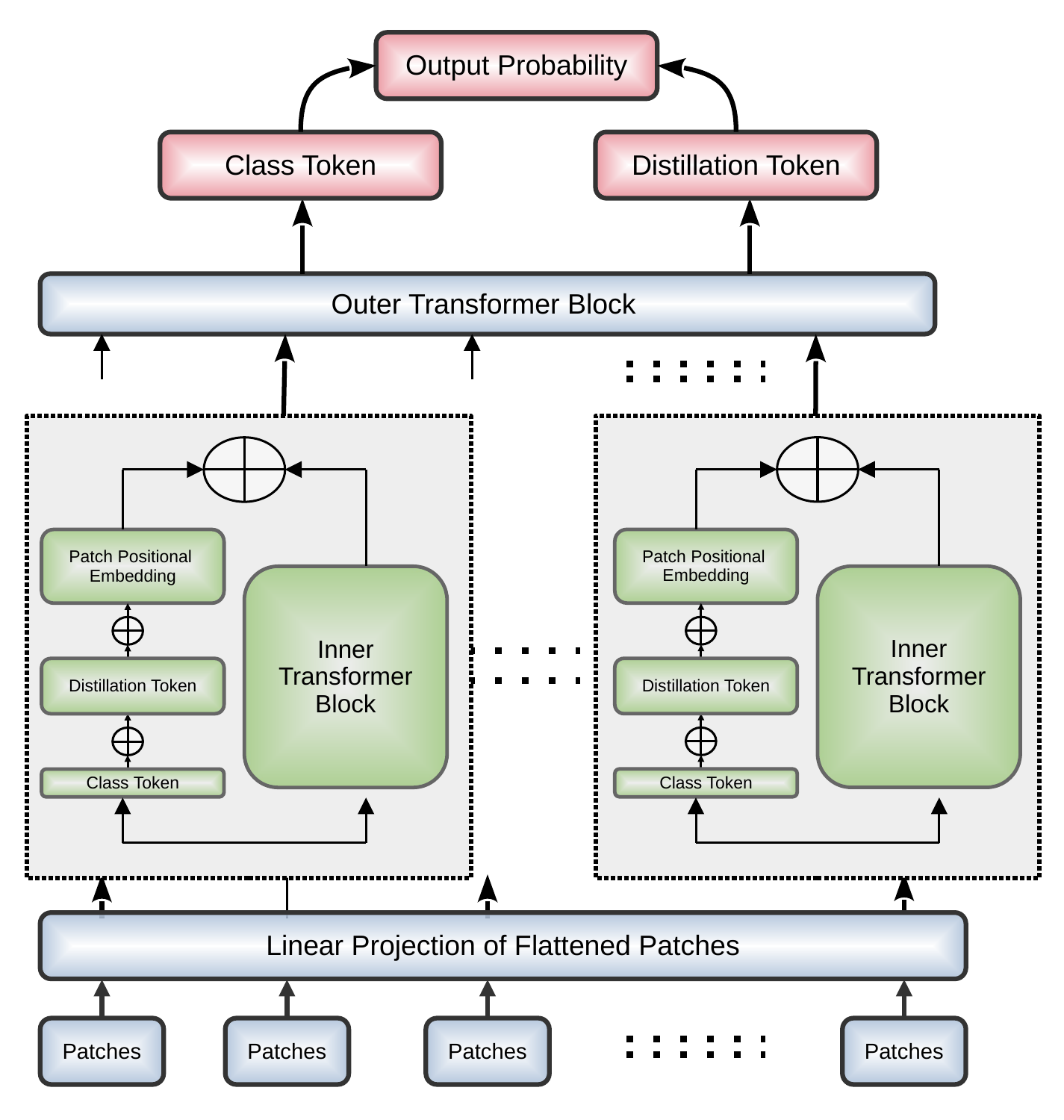}
    \caption{Neural architecture of the proposed \textsc{TITN}. The converted image patches have been fed into the inner transformer block, then the outer transformer block, and finally processed into class and distillation tokens.}
    \label{fig:titn_module}
\end{figure}

Figure \ref{fig:titn_module} presents an illustration of the suggested network. The inner-transformer-block (figure \ref{fig:transformer_block}) in this architecture extracts local features, whereas the outer-transformer-block (figure \ref{fig:transformer_block}) extracts global features. We create the full network by stacking the entire block up to the number of depths.

\begin{algorithm}[b!]
\caption{Pseudo-code of the Proposed TITN}
\begin{algorithmic}[1]
\STATE \textbf{Input:} Image Patches $\{p_1, p_2, \ldots, p_n\}$
\STATE $\mathcal{P} \gets \text{LinearProjection}(\{p_1, p_2, \ldots, p_n\})$
\FOR{$p \in \mathcal{P}$}
    \STATE $h_p \gets \text{InnerTransformerBlock}(p, \theta_{\text{inner}})$
\ENDFOR
\STATE $H \gets \{h_p \mid p \in \mathcal{P}\}$
\STATE $O \gets \text{OuterTransformerBlock}(H, \theta_{\text{outer}})$
\STATE $y_{\text{class}} \gets \text{ClassToken}(O)$
\STATE $y_{\text{distill}} \gets \text{DistillationToken}(O)$
\STATE $y_{\text{output}} \gets \text{Concat}(y_{\text{class}}, y_{\text{distill}})$ 
\STATE $\hat{y} \gets \text{OutputProbability}(y_{\text{output}})$
\STATE \textbf{Output:} $\hat{y}$
\end{algorithmic}
\end{algorithm}

The inner MLP block receives input after that, and the output is added to itself. After that, the generated patch is rearranged and linearly scaled to a larger patch size. The first and last rows, for the class token and distillation token, respectively, are then zero-padded. These are then referred to as patch residuals \cite{cho2019efficacy}. Then, the larger patches are added together with the patch residuals. The output from these last, larger patches is added to the output from the outer attention blocks. The outer MLP block receives this output, which is then combined with the input to produce the final output of our design. From the final output, the classification token and distillation token are sliced and projected using a fully connected layer \cite{park2019relational}. As, this end-to-end network pipeline is specifically built for image classification purposes, we did not add any decoder to the current architecture. However, for other tasks, such as image super-resolution, image generation etc. decoder will be added.

\subsection{Datasets}
To evaluate the effectiveness of our proposed \textsc{TITN} model, we have chosen three benchmark datasets for conducting various experiments. These datasets cover different domains and characteristics, allowing us to compare our model with other state-of-the-art methods.

\subsubsection{MNIST}
MNIST is a straightforward real-world data set that doesn't require a lot of preparation or formatting. There are 10,000 examples in the test set and 60,000 examples in the training set, for a total of 70,000 handwritten digit greyscale images. The photos have been centered in a $28 \times 28$ grid and normalized. There are a total of 10 separate classes, each representing a number from 0 to 9. The training dataset includes labels to show the model what each digit should look like. The model is then tested using the test dataset, which is fed only photos to allow it to forecast data that it has never seen before.

\subsubsection{CIFAR10}
The CIFAR-10 dataset (Canadian Institute for Advanced Research, 10 classes), which comprises 60,000 photos and contains 50,000 training images and 10,000 test images, is a subset of the 80 million-image Tiny Images dataset. The RGB images are composed of $32 \times 32$ pixels. The images are divided into ten categories that are all mutually exclusive: truck, ship, frog, horse, airplane, car (but not pickup truck or truck), cat, deer, and dog (but not pickup truck). Each class comprises exactly 6,000 images with the main theme belonging to that category.

\subsubsection{CIFAR100}
The CIFAR-100 dataset (Canadian Institute for Advanced Research, 100 classes) is a subset of the Tiny Images dataset and consists of 60,000, $32 \times 32$, color images. The CIFAR-100's 100 classes are divided into 20 super-classes. Per class, there are 100 testing images.and 500 training images. Each image has a "fine" and a "coarse" designation, indicating the class to which it belongs (the superclass to which it belongs).

\subsection{Augmentations}
In order to expose the model to a greater variety of training examples, picture augmentation involves applying changes to photos to produce different versions of the same material. For our goals, we applied Auto Augment~\cite{cubuk2019autoaugment}, CutMix~\cite{yun2019cutmix}, Random Crop, and Random Horizontal Flip augmentations to our training dataset before converting them to tensors and normalizing them.

\subsection{Loss Function}
Our proposed loss function leverages both distillation loss and cutmix augmentation loss in order to accurately classify image data. We will go through the loss function we utilized for our network structure in this section.

\subsubsection{Cross Entropy Loss Function(CE)}
The effectiveness of a classification model whose output is a probability value between 0 and 1 is measured by cross-entropy loss, also known as log loss. As the anticipated probability departs from the labelled probability, cross-entropy loss grows. In binary classification, where the number of classes $M=2$, Binary Cross-Entropy(BCE) can be calculated as \cite{martinez2018taming}:
\begin{equation}
    \mathcal{L}_\mathbf{{BCE}}(p, l) = -{(y\log(p) + (1 - y)\log(1 - p))}
\end{equation}
If $M > 2$ (i.e. multiclass classification), For each class label per observation, we compute a separate loss and then add the results.
\begin{equation}
    \mathcal{L}_{\textbf{CE}}(p, l) = -\sum_{c=1}^My_{o,c}\log(p_{o,c})
\end{equation}
where p, l are the prediction and label, o is the sample index and c is the class index.


\subsubsection{CutMix Loss function(CM)} 
CutMix offers several advantages over other augmentation methods: it enhances model robustness by combining patches of different images and labels, promotes localized data augmentation, and mitigates the risk of overfitting. It also improves performance in tasks like image classification, object detection, and adversarial robustness by effectively utilizing both regional and global information from the images. Using the CutMix augmentation, which replaces an image region with a patch from another training image at a random shape, the training dataset is augmented. The shape of the replacement patch is determined by the value of lambda ($\lambda$). The value of alpha, 0.5 for our case, is used to sample lambda($\lambda$) from a beta distribution. As a result, there are two labels on the training images. So, the following function is utilized to calculate the loss.
\begin{equation}
    \mathcal{L}_{CM}(p, l_1, l_2) = \lambda*\mathcal{L}_{CE}(p, l_1) + (1-\lambda)*\mathcal{L}_{CE}(p, l_2)
\end{equation}
where, p is the output prediction, $l_1$, $l_2$ are  label-1 and label-2 respectively.

\begin{figure}[t!]
    \centering
    \includegraphics[width = 0.50\textwidth]{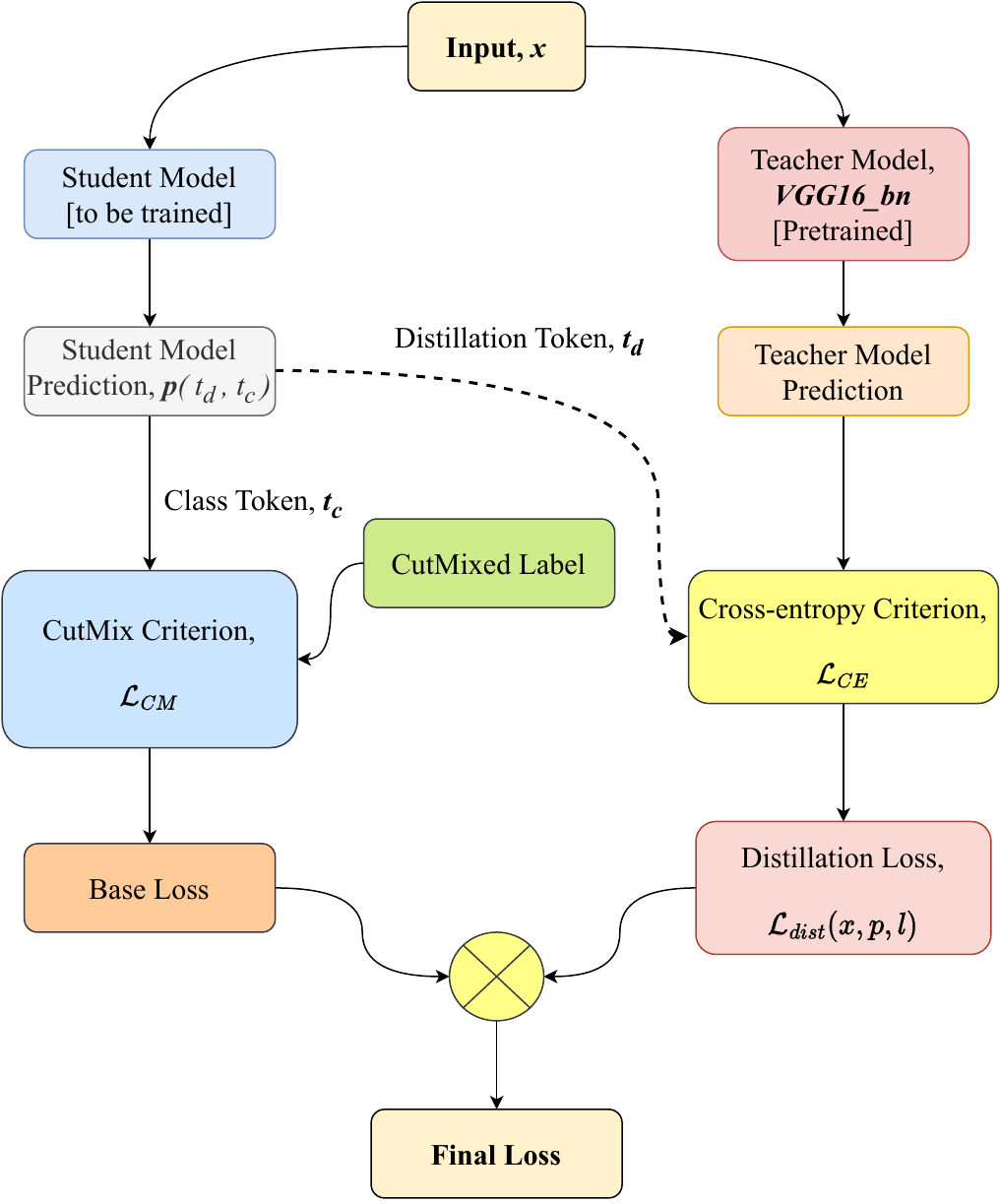}
    \caption{Proposed loss function leveraging both distillation loss and cross-entropy loss. The loss function utilizes both cutmix and cross-entropy criteria for student and teacher models respectively.}
    \label{fig:loss_function}
\end{figure}

\begin{algorithm}[htbp]
\caption{Pseudo-code of the Proposed Loss Function}
\begin{algorithmic}[1]
\STATE \textbf{Input:} $x$
\STATE $p_{\theta}(y_t | x, t) \gets \text{StudentModel}(x)$
\STATE $T(y_t | x, \theta_T) \gets \text{TeacherModel}(x)$
\STATE $t_d \gets \text{DistillationToken}(p_{\theta}(y_t | x, t), T(y_t | x, \theta_T))$
\STATE $y_c \gets \text{CutMixedLabel}(t_d, t)$
\STATE $\mathcal{L}_{\text{CE}} \gets \text{CrossEntropyLoss}(p_{\theta}(y_t | x, t), T(y_t | x, \theta_T))$
\STATE $\mathcal{L}_{\text{CutMix}} \gets \text{CutMixCriterion}(p_{\theta}(y_t | x, t), y_c)$
\STATE $\mathcal{L}_{\text{Base}} \gets \text{BaseLoss}(\mathcal{L}_{\text{CE}}, \mathcal{L}_{\text{CutMix}})$
\STATE $\mathcal{L}_{\text{Distill}}(\theta, \theta_T) \gets \text{DistillationLoss}(\mathcal{L}_{\text{Base}}, \theta, \theta_T)$
\STATE $\mathcal{L}_{\text{Final}} \gets \mathcal{L}_{\text{Base}} + \mathcal{L}_{\text{Distill}}(\theta, \theta_T)$
\STATE \textbf{Output:} $\mathcal{L}_{\text{Final}}$
\end{algorithmic}
\end{algorithm}

\subsubsection{Distillation Loss function} 
Our custom loss function (figure \ref{fig:loss_function}), which is a cross between the CutMix and Cross-entropy loss functions, accepts inputs and outputs the ultimate loss value. Training image ($x$), prediction by the student model ($p$), and label($l$) are essentially the three inputs. The classification token ($t_c$) and distillation token ($t_d$) make up the tuple($p$) that the student model predicts. A prediction is made using the input($x$) by the pre-trained teacher model. We obtain the instructor label($l_t$) by selecting the argument that yields the maximum prediction.  The following equation is used to determine the final loss:

\begin{equation}
    \mathcal{L}_{dist}(x, p, l) = \alpha*L_{CM}(t_c, l) + (1-\alpha)*\mathcal{L}_{CE}(t_d, l_t)
\end{equation}

where the value for $\alpha$ in this instance is 0.5.

\subsection{Distillation Process}
We have included the distillation token, a learnable parameter with a randomized initialization that interacts with the patch tokens and classification tokens via the multi-head self-attention layers. After the last layer, the network produced this token along with the classification token. Similar to a classification token, it is learned using back-propagation. With one exception, the distillation tokens' goal when creating the output is to duplicate the teacher's projected soft label rather than the actual label. While still complementary to the class embedding, the distillation token enables our model to learn from the teacher's output. The small student model mimics the teacher by applying the soft label, which leads to quicker convergence and greater performance than other models.

\section{Results \& Discussion}
In this section, the training settings and the results obtained from experiments have been discussed. Furthermore, for a better understanding of the assessments for the proposed \textsc{TITN}, a brief introduction of the datasets utilized has been provided.

\begin{table}[hb!]
   \centering
   \caption{Hyper-parameter Settings Used for Different Student Models}
   \label{tab:student-params}
   \resizebox{0.5\textwidth}{!}{%
   \begin{tabular}{cccccc}
       \toprule
       \textbf{Parameter} & \textbf{ViT} & \textbf{DeiT} & \textbf{TNT} & \textbf{TITN} & \textbf{TITN(Large Patch)} \\
       \midrule
       Image Size & 32$\times$32 & 32$\times$32 & 32$\times$32 & 32$\times$32 & 32$\times$32 \\
       Patch Dimension & 192 & 192 & 192 & 192 & 192\\
       Pixel Dimension & - & - & 12 & 12 & 12\\
       Patch Size & 8 & 8 & 8 & 8 & 16 \\
       Pixel Size & - & - & 2 & 2 & 4\\
       Depth & 12 & 12 & 12 & 12 & 12 \\
       Parameter Count & 5.36M & 5.37M & 5.83M & 5.85M & 5.85M\\
       \bottomrule
   \end{tabular}
   }
\end{table}

\begin{figure}[t!]
    \centering
    \includegraphics[width=0.6\textwidth]{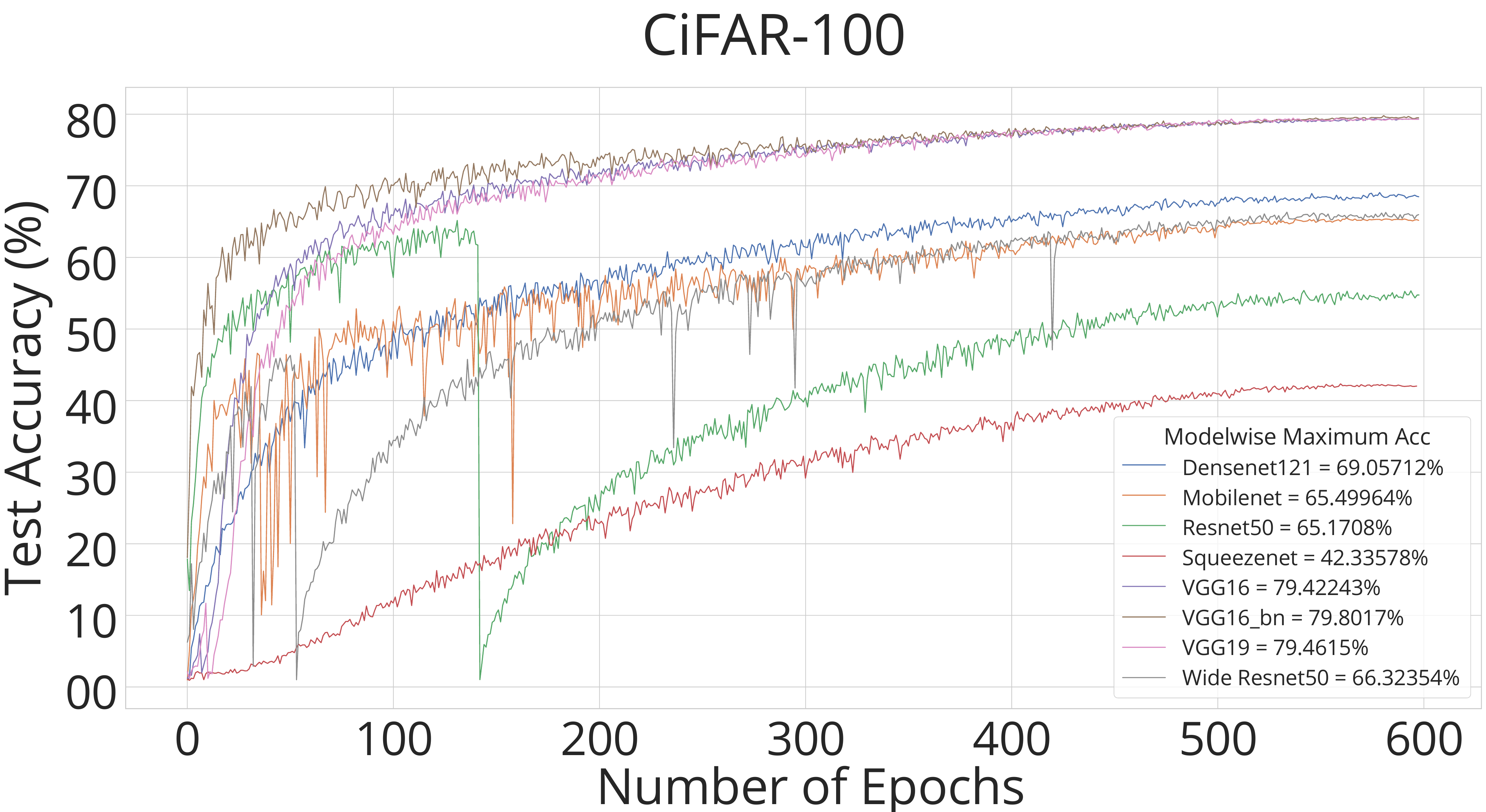}
    \caption{Performance comparison on CIFAR-100 dataset against various pre-trained teacher models.}
    \label{fig:Cifar100_teachermodel_comparison}
\end{figure}

\begin{figure}[t!]
    \centering
    \includegraphics[width=0.6\textwidth]{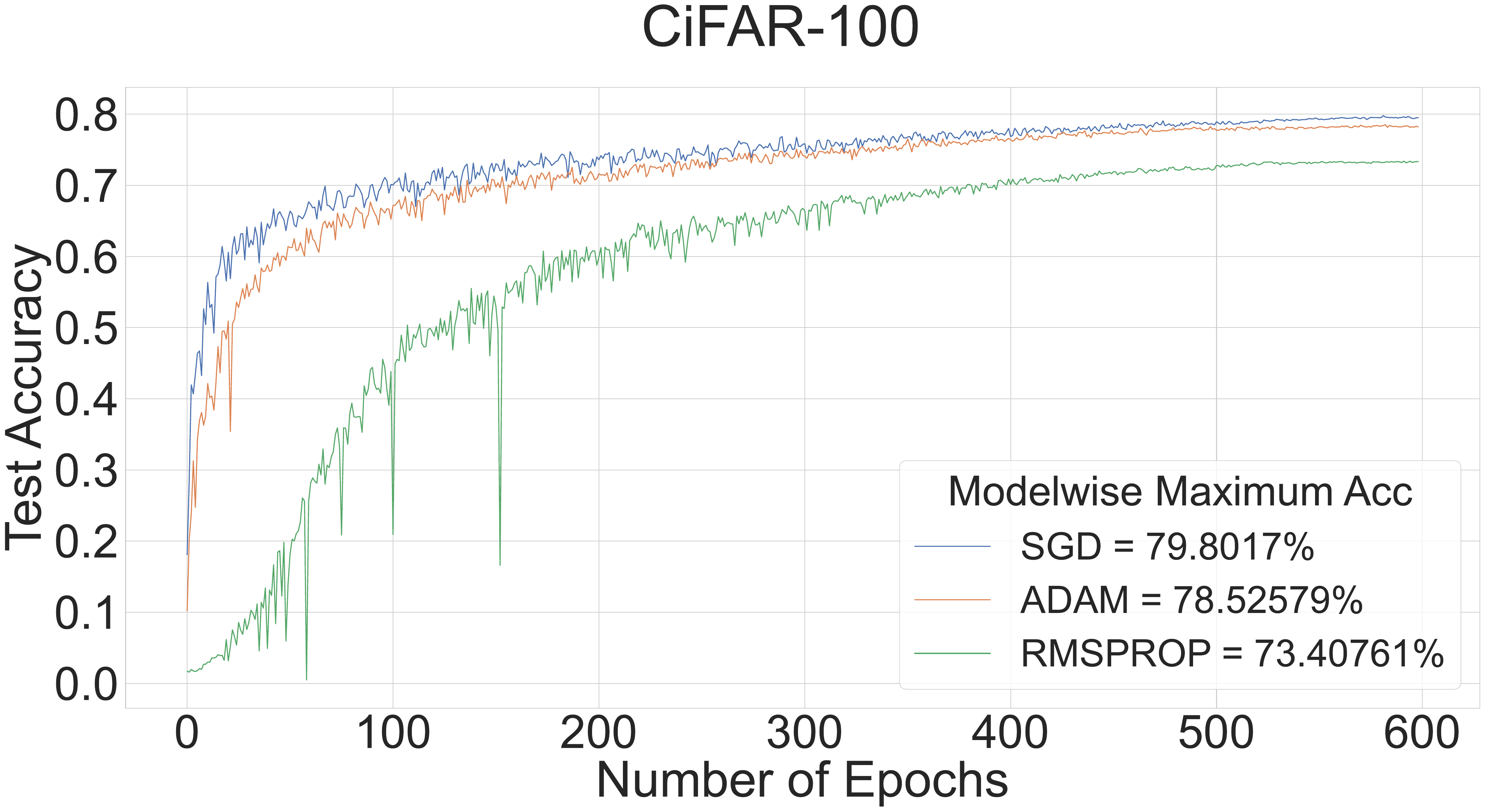}
    \caption{Performance evaluation on CIFAR-100 using VGG16-BN as teacher model.}
    \label{fig:Cifar100_teachermodel_optimizer_comparison}
\end{figure}

\begin{table}[hb!]
\centering
\caption{Performance Evaluation on CIFAR-100 Dataset for Selecting Best Teacher Model}
\label{tab:teacherc100}
\resizebox{0.5\textwidth}{!}{
    \begin{tabular}{ccc}
    \toprule
    \textbf{Teacher Models} & \textbf{Top-1 Accuracy (\%)} & \textbf{Top-5 Accuracy (\%)} \\
    \midrule
    Densenet121 & 69.06 & 89.61 \\
    Mobilenet\_v3 & 65.50 & 88.86 \\
    Resnet50 & 65.17 & 87.32 \\
    WideResnet50 & 66.32 & 87.91 \\
    Squeezenet & 42.34 & 67.13 \\
    VGG16 & 79.42 & 94.72 \\
    \textbf{VGG16\_bn} & 79.80 & 94.51 \\
    VGG19 & 79.46 & 94.75 \\
    \bottomrule
    \end{tabular}
    }
\end{table}

\subsection{Experimental Settings}
The Pytorch framework and Python 3.8 have been used for all of the experiments. We used a single GPU with 16GB of RAM. The batch size is 1024. We have regularized the dataset for the CIFAR-10 and CIFAR-100 using Auto-Augment and CutMix augmentation, as well as other transformations like randomcrop and randomhorizontalflip. We have adopted a different strategy for the MNIST dataset, in which the image is regularized using a variety of transformations, including resize, random crop, random rotation, random affine, and collerjitter. The images from these various datasets were all then normalized. We used the SGD optimizer with 0.9 momentum and 1e-4 weight decay. Using the Cosine Annealing LR scheduler, the initial learning rate of 0.1 was gradually reduced.

Here, ViT, DeiT, TNT, and our TITN are four different transformer-based student models that we have trained. Table \ref{tab:student-params} comprises the settings for these models.

\begin{table*}[t]
   \centering
   \caption{Performance evaluation on different SOTA transformer-based models against our proposed \textbf{TITN} using the CIFAR-100 dataset.}
   \resizebox{0.90\textwidth}{!}{%
   \begin{tabular}{@{}l l c c c c c c c@{}}
       \toprule
       & \textbf{Model Name} & \textbf{Top-1 Acc (\%)} & \textbf{Top-5 Acc (\%)} & \textbf{Precision} & \textbf{Recall} & \textbf{F1-Score} & \textbf{GFLOPs} & \textbf{Parameters (M)} \\
       \midrule
       \textbf{Teacher Model} & VGG16\_bn & 79.80 & 94.51 & 0.77 & 0.80 & 0.79 & 448.47 & 138.36 \\
       \midrule
       \multirow{6}{*}{\textbf{Student Models}} 
       & ViT & 70.79 & 89.55 & 0.65 & 0.71 & 0.68 & 93.37 & 5.39 \\
       & DeiT & 73.91 & 91.91 & 0.71 & 0.74 & 0.73 & 98.84 & 5.41 \\
       & TNT & 68.47 & 88.44 & 0.61 & 0.68 & 0.64 & 105.93 & 5.83 \\
       & \textbf{TITN} & \textbf{74.71} & \textbf{92.28} & \textbf{0.72} & \textbf{0.75} & \textbf{0.74} & 111.40 & 5.85 \\
       & TITN (MixUp) & 74.40 & 92.53 & 0.70 & 0.74 & 0.72 & 111.40 & 5.85 \\
       & TITN (Large Patch) & 63.63 & 85.93 & 0.58 & 0.64 & 0.61 & 36.09 & 5.85 \\
       \bottomrule
   \end{tabular}
   }
   \label{cif100tab}
\end{table*}

\begin{figure}[t]
    \centering
    \includegraphics[width = 0.6\textwidth]{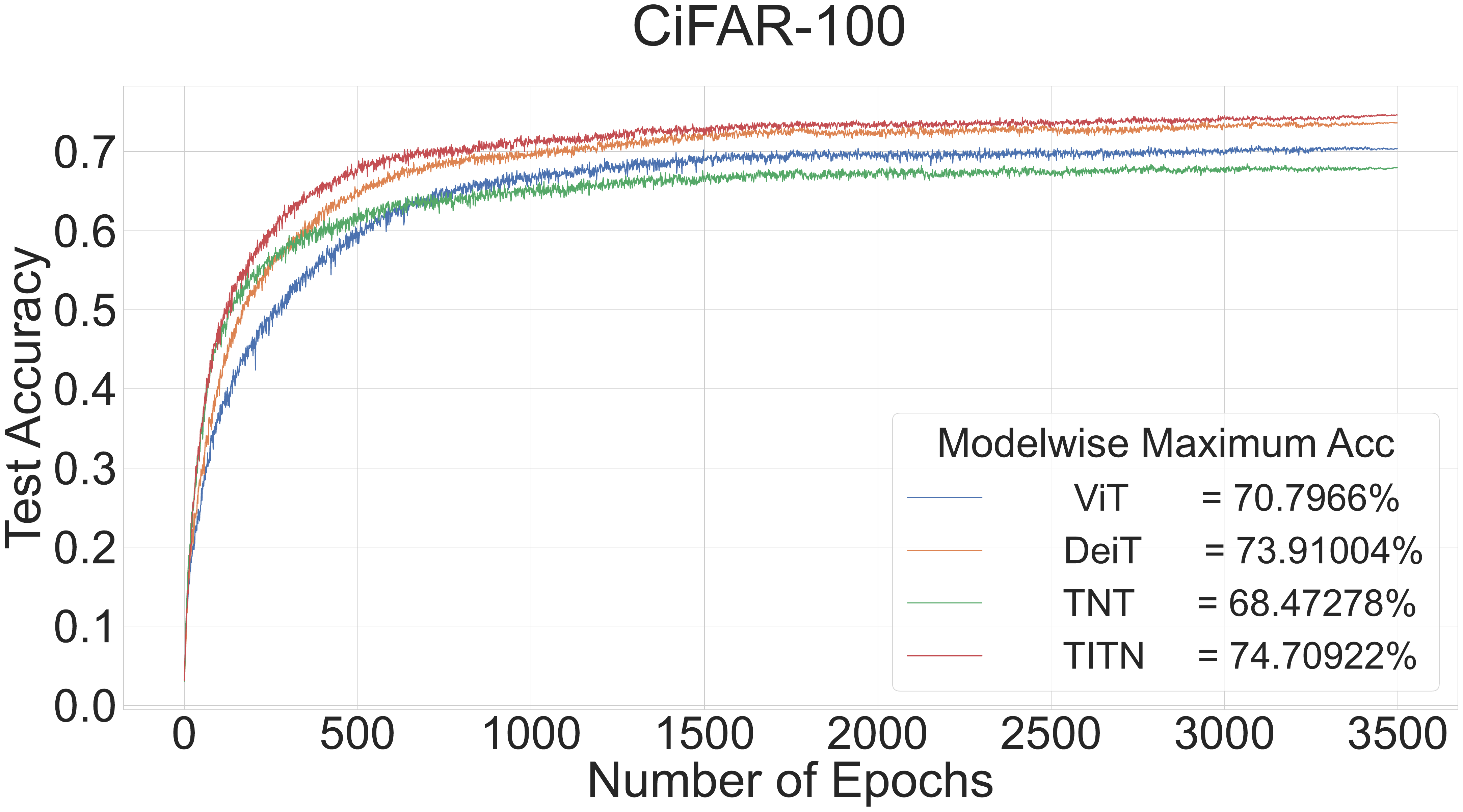}
    \caption{Performance evaluation of our model against various SOTA models using CIFAR-100 dataset. }
    \label{fig:cifar_100_top_1}
\end{figure}

\subsection{Results}
This subsection primarily comprises performance demonstration of various teacher and student models based on their maximum test accuracy, top-1 accuracy and top-5 accuracy.

\par
\textbf{Evaluation of Teacher Model:} In order to facilitate effective learning among student models, the selection of a proficient teacher model was imperative. To assess the efficacy of teacher models, their performance was evaluated using the CIFAR-100 dataset. CIFAR-100, characterized by its complexity owing to a limited volume of data but a substantial number of distinct classes, served as an ideal benchmark dataset for discerning the superior-performing teacher models.
 
Table \ref{tab:teacherc100} summarized the performance of eight Teacher models based on their top-1 and top-5 accuracy obtained from experiments on the CIFAR-100 dataset. An intriguing observation that emerged was that, while VGG16\_bn exhibited the highest top-1 accuracy of 79.80\%, the highest top-5 accuracy was found in the case of VGG19 with 94.75\%. On the contrary, Squeezenet demonstrated both the lowest top-1 accuracy of 42.34\% and the lowest top-5 accuracy of 67.13\%. The teacher model with the highest performance, VGG16\_bn, was further evaluated by employing three different optimizers - SGD, ADAM, and RMSPROP - to investigate any fluctuation in maximum test accuracy. The outcomes were visually depicted in Figure \ref{fig:Cifar100_teachermodel_optimizer_comparison}, which illustrated that the highest test accuracy was obtained by SGD at 79.8017\% and the lowest was obtained by RMSPROP at 73.40761\%. The SGD-optimized test accuracy of the models was plotted against 600 epochs, as illustrated in Figure \ref{fig:Cifar100_teachermodel_comparison}. Based on the graph, it was observed that VGG16\_bn demonstrated the highest test accuracy, while the lowest test accuracy was found to belong to Squeezenet at only 42.336\%. It was also noted that the accuracy curve for VGG16\_bn converged faster than that of the other models.

\begin{table}[t]
   \centering
   \caption{\textbf{CIFAR-10} Dataset Performance Evaluation on Different SOTA transformer-based models against Our Proposed \textbf{TITN}}
   \resizebox{0.6\textwidth}{!}{%
   \begin{tabular}{cccc}
        \toprule
        &\textbf{Model Name} &  \textbf{Top-1 Acc(\%)} & \textbf{Top-5 Acc(\%)} \\
        \midrule
        Teacher Model & \textbf{VGG16\_bn}\cite{simonyan2014very} & 95.98 & 99.92 \\
        \addlinespace
        \midrule
        \multirow{4}{*}{Student Models} & ViT & 86.33 & 99.41 \\
        & Deit & 91.64 & 99.73 \\
        & TNT & 86.91 & 99.38 \\
        & \textbf{TITN} & 92.03 & 99.8 \\
        & TITN(Large Patch)  & 85.39 &  99.31 \\ 
        \bottomrule
    \end{tabular}
   }
    \label{cif10}
\end{table}

    
    

   

\textbf{Results on CIFAR-100:} As demonstrated in table \ref{cif100tab}, the teacher model, VGG16\_bn, exhibited a top-1 accuracy of 79.80\% and a top-5 accuracy of 94.51\%, with precision, recall, and F1-score of 0.77, 0.80, and 0.79, respectively. In contrast, the student models showed varying performance metrics. Among them, the proposed TITN model attained the highest top-1 accuracy of 74.71\% and a top-5 accuracy of 92.28\%, with precision, recall, and F1-score of 0.72, 0.75, and 0.74, respectively. ViT achieved a top-1 accuracy of 70.79\% and a top-5 accuracy of 89.55\% while ViT demonstrated a top-1 accuracy of 70.79\% and a top-5 accuracy of 89.55\%. TNT achieved a top-1 accuracy of 68.47\% and a top-5 accuracy of 88.44\%.  Additionally, the TITN (MixUp) model showed a top-1 accuracy of 74.40\% and the highest top-5 accuracy of 92.53\%, whereas the TITN (Large Patch) model exhibited the lowest performance with a top-1 accuracy of 63.63\% and a top-5 accuracy of 85.93\%. The computational complexities of the models were also reported in terms of GFLOPs and the number of parameters, with TITN generic and MixUp having the highest computational cost and parameter count among the student models.


\begin{figure}[t]
    \centering
    \includegraphics[width = 0.6\textwidth]{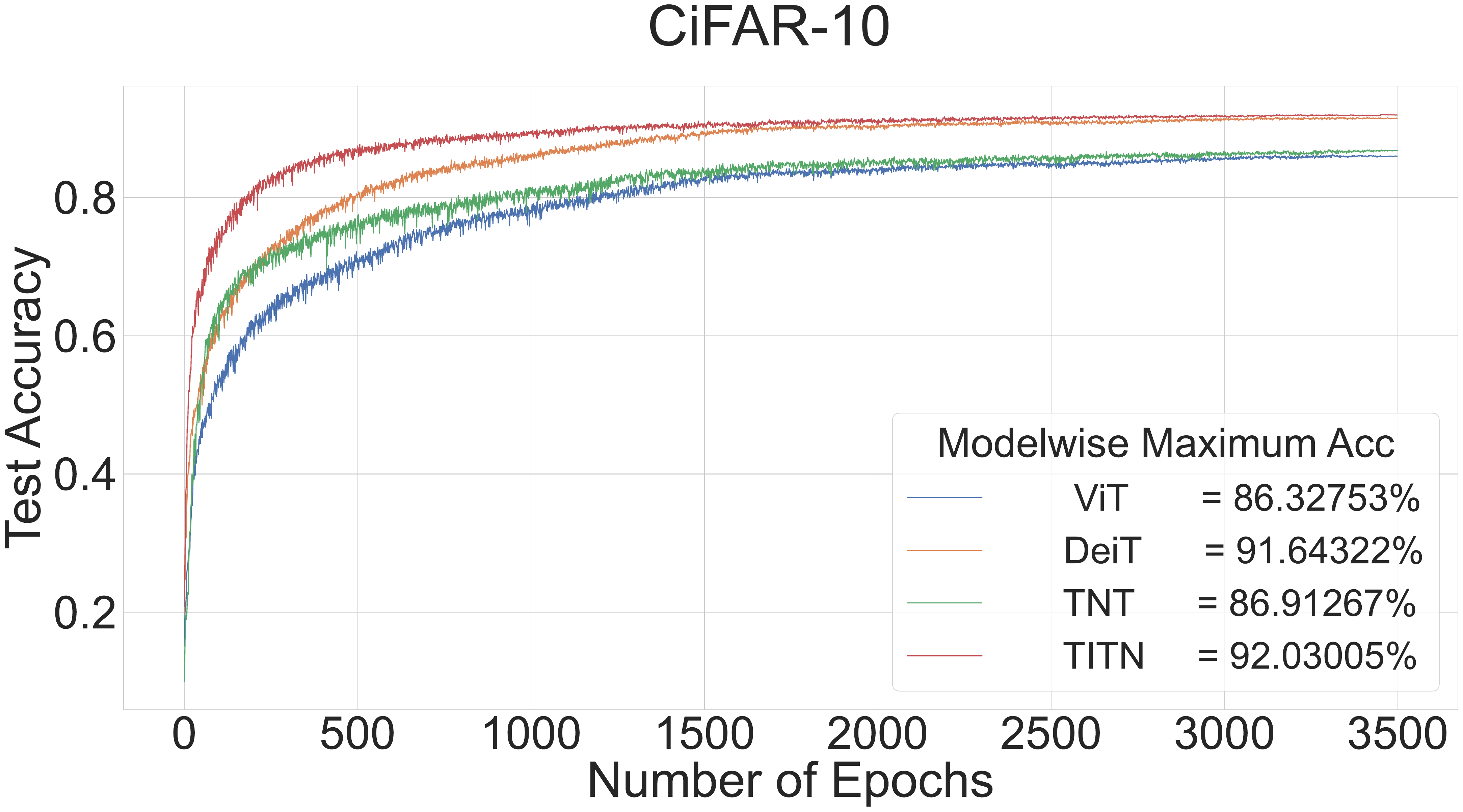}
    \caption{Top 1\% accuracy comparison on CIFAR-10 dataset}
    \label{fig:cifar_10_top_1}
\end{figure}

\textbf{Results on CIFAR-10:} As delineated by table \ref{cif10}, the teacher model VGG16\_bn achieved top-1 and top-5 accuracy at 95.98\% and 99.92\%, respectively. Among the student models, our proposed TITN model demonstrated superior performance with a top-1 accuracy of 92.03\% and a top-5 accuracy of 99.80\%, closely rivalling the teacher model.
The DeiT model also performed well, with a top-1 accuracy of 91.64\% and a top-5 accuracy of 99.73\%. Other student models, including ViT and TNT, showed competitive performance but did not match the accuracy of our TITN model. The TITN (Large Patch) variant, while showing good performance, had a lower top-1 accuracy of 85.39\% and a top-5 accuracy of  99.31\%, indicating a balance between computational efficiency and accuracy.

\begin{figure}[t!]
    \centering
    \includegraphics[width = 0.6\textwidth]{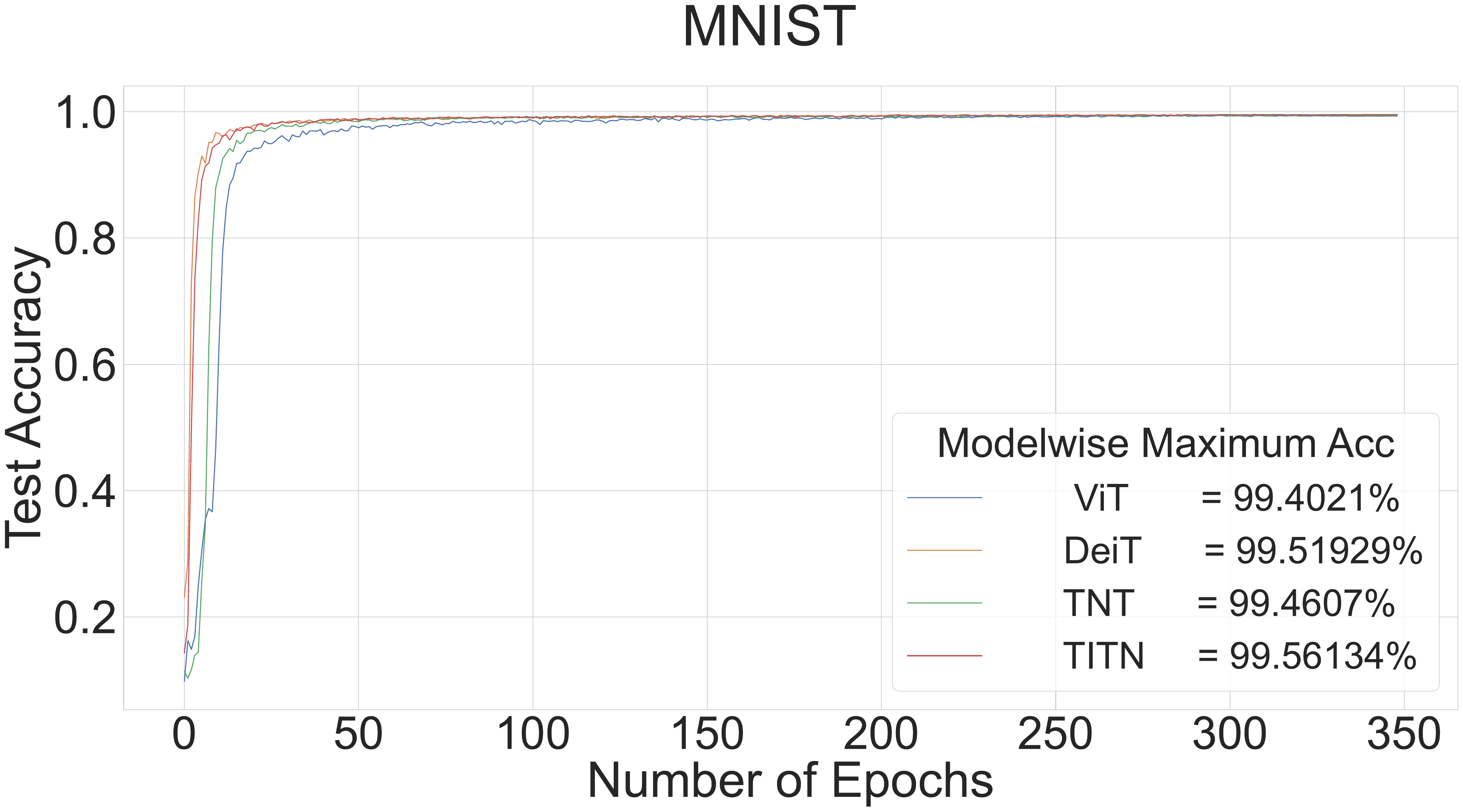}
    \caption{Performance evaluation of our model against SOTA model using MNIST dataset.}
    \label{fig:mnist_top_1}
\end{figure}

\textbf{Results on MNIST:} As evident from Fig. \ref{fig:mnist_top_1}, all four of the student models had almost the same top-1 accuracy for the MNIST dataset. Still, the proposed TITN model performed slightly better than the other with a maximum top-1 test accuracy of 99.56\% after 350 epochs which was 0.1\% higher than ViT (99.4\%), 0.04\% higher than Distill-ViT (99.52\%) and 0.1\% higher than the TNT model (99.46\%) due to robust representation learning capabilities. To sum it up, table \ref{mnisttab} demonstrates that for the MNIST dataset, the teacher model, VGG16\_bn model had a top-1 accuracy of 99.75\% and a top-5 accuracy of 100\%. The proposed student model TITN exhibited the highest top-1 accuracy of 99.56\%. All the student models for this data set had a top-5 accuracy of 100\%.

\subsection{Baseline Comparative Results}
The comparative analysis of knowledge distillation results in table \ref{tab:kd_results} reveals significant variations in top-1 accuracy drop across different models. Notably, our approach (VGG16-bn to TITN) with Cross-Arch. Cutmix demonstrates a minimal accuracy reduction of 3.93\%, from 95.98\% to 92.05\%. In contrast, the Attn. Probe method for distilling DeiT to DeiT Tiny on CIFAR-10 sees a 5.1\% decrease, highlighting the robustness of our method. Cross-Arch. distillation from ViT-B to ResNet50 results in a 2.63\% accuracy drop, while Swin-L to ResNet50 shows an 11.04\% reduction, indicating significant performance losses. Similarly, Swin-Tiny to EfficientNet-B0 and Swin-L to MobileNetV2 experience accuracy drops of 1.8\% and 5.16\% respectively. Our approach outperforms these transformer-based methods in maintaining higher accuracy with a relatively smaller model size and computational cost.

\begin{table}[t!]
    \centering
    \caption{Performance evaluation on different SOTA transformer-based models against our proposed \textbf{TITN} using MNIST dataset}
    \label{mnisttab}
    \resizebox{0.6\textwidth}{!}{%
    \begin{tabular}{@{}lccc@{}}
        \toprule
        & \textbf{Model Name} & \textbf{Top-1 Acc(\%)} & \textbf{Top-5 Acc(\%)} \\
        \midrule
        Teacher Model & \textbf{VGG16\_bn} & 99.75 & 100 \\
        \addlinespace
        \midrule
        \multirow{4}{*}{Student Models} & ViT & 99.40 & 100 \\
        & DeiT & 99.52 & 100 \\
        & TNT & 99.46 & 100 \\
        & \textbf{TITN} & 99.56 & 100 \\
        \bottomrule
    \end{tabular}
    }
\end{table}

\subsection{Discussion}


The experimental findings unambiguously show the effectiveness of our strategy. Across all datasets studied, the TITN model consistently outperforms previous state-of-the-art approaches, setting new accuracy benchmarks. The outstanding results on the CIFAR-100, CIFAR-10, and MNIST datasets, where the proposed model achieves unparalleled top-1 and top-5 accuracy rates, are particularly noteworthy. These achievements validate our architecture's capacity to holistically collect local and global picture properties, contributing to improved classification accuracy. The intrinsic challenges of training transformer models, including their resource, time, and data requirements, are recognized. The incorporation of knowledge distillation overcomes these issues by facilitating efficient learning via a student-teacher paradigm. The distillation process allows the student model to benefit from the teacher's expertise by exploiting the insights of a bigger teacher model, resulting in enhanced convergence speed and accuracy. This strategy is especially important given the growing demand for resource-efficient deep learning models. The TITN architecture's success is intimately linked to the concept of knowledge distillation. The capacity of the student model to duplicate the teacher's output, helped by soft labels generated by the distillation token, is a critical aspect in achieving greater performance. This method not only allows the student model to learn from the teacher's accumulated knowledge, but it also helps to reduce overfitting and promote generalization to previously unseen data. The knowledge distillation technique efficiently conveys to the student the essence of the broader teacher model, overcoming the restrictions associated with limited training data.

\begin{table}[ht!]
   \centering
   \caption{Comparative knowledge distillation results for various teacher-student model pairs. FLOPs and parameter counts are provided for each model configuration.}
   \resizebox{\textwidth}{!}{
   \begin{tabular}{@{}lccccccc@{}} 
       \toprule
       \multirow{2}{*}{\textbf{Teacher $\rightarrow$ Student}} & \multirow{2}{*}{\textbf{KD Approach}} & \multicolumn{2}{c}{\textbf{Top-1 Accuracy (\%)}} & \multirow{2}{*}{\textbf{FLOPs (G)}} & \multicolumn{2}{c}{\textbf{Parameters (M)}} \\
       \cmidrule(lr){3-4} \cmidrule(lr){6-7}
       & & \textbf{Before} & \textbf{After} & & \textbf{Before} & \textbf{After} \\
       \midrule
       ResNet110 $\rightarrow$ ResNet20 \cite{lin2022knowledge} & TaT & 74.31 & 71.70 (-2.61$\downarrow$) & 0.255 & 1.7 & 0.27 \\
       ResNet56 $\rightarrow$ ResNet20 \cite{lin2022knowledge} & TaT & 72.00 & 70.06 (-2.06$\downarrow$) & 0.2 & 0.85 & 0.27 \\
       ResNet110 $\rightarrow$ ResNet32 \cite{lin2022knowledge} & TaT & 74.31 & 73.08 (-1.23$\downarrow$) & 0.25 & 1.7 & 0.46 \\
       DeiT $\rightarrow$ DeiT Tiny (CIFAR-10) \cite{zhan2022stage} & Attn. Probe & 76.30 & 71.82 (-5.10$\downarrow$) & 1.38 & 21.3 & 2.38 \\
       ViT-B $\rightarrow$ ResNet50 \cite{liu2022cross} & Cross-Arch. & 90.02 & 87.39 (-2.63$\downarrow$) & 55.4 & 86 & 25.4 \\
       Swin-L $\rightarrow$ ResNet50 \cite{liu2022cross} & Cross-Arch. & 87.32 & 76.28 (-11.04$\downarrow$) & 103.9 & 197 & 25.4 \\
       Swin-Tiny $\rightarrow$ EfficientNet-B0 \cite{liu2022cross} & Cross-Arch. & 94.50 & 92.70 (-1.80$\downarrow$) & 5.3 & 5.3 & 4.7 \\
       Swin-L $\rightarrow$ MobileNetV2 \cite{liu2022cross} & Cross-Arch. & 93.50 & 88.34 (-5.16$\downarrow$) & 103.9 & 197 & 6.0 \\
       \textbf{VGG16-BN $\rightarrow$ TITN (Ours)} & Cross-Arch. + CutMix & 95.98 & 92.05 (-3.93$\downarrow$) & 444.32 & 134.30 & 5.8 \\
       \bottomrule
   \end{tabular}
   }
   \label{tab:kd_results}
\end{table}

\section{Conclusion}
This paper introduces a network that ingeniously merges a transformer-in-transformer architecture with knowledge distillation, thereby adeptly mitigating these limitations. The study accentuates the reciprocal relationship between pioneering concepts and pragmatic applicability. Empirical findings underscore the superiority of the proposed model in terms of both execution speed and precision, when contrasted with established models such as ViT, DeiT, and TNT, across various datasets. This study establishes a performance yardstick that effectively confronts the complexities of execution swiftness and precision. Though our solution has shown significant performance in visual recognition, there is still room for improvement from a computational perspective. Our student-teacher model learns faster due to our proposed loss criterion, however, the model is computationally larger in comparison to other methods. Future work can be done to reduce the computational usage. Looking ahead, the research advocates for the integration of a novel attention mechanism into the foundational transformer architecture. This envisioned augmentation will hold the promise of substantially amplifying an array of comprehensive performance metrics.

\bibliographystyle{IEEEtran}
\bibliography{reference}

\end{document}